# SAFETY ASSESSMENT OF SCAFFOLDING ON CONSTRUCTION SITE USING AI

Sameer Prabhu[1], Amit Patwardhan[1], and Ramin Karim[1]

[1] Luleå university of technology, Luleå, 97187 Sweden

[1] *firstname.lastname@ltu.se*

## ABSTRACT

In the construction industry, safety assessment is vital to ensure both the reliability of assets and the safety of workers. Scaffolding, a key structural support asset requires regular inspection to detect and identify alterations from the design rules that may compromise the integrity and stability. At present, inspections are primarily visual and are conducted by site manager or accredited personnel to identify deviations. However, visual inspection is time-intensive and can be susceptible to human errors, which can lead to unsafe conditions. This paper explores the use of Artificial Intelligence (AI) and digitization to enhance the accuracy of scaffolding inspection and contribute to the safety improvement. A cloud-based AI platform is developed to process and analyse the point cloud data of scaffolding structure. The proposed system detects structural modifications through comparison and evaluation of certified reference data with the recent point cloud data. This approach may enable automated monitoring of scaffolding, reducing the time and effort required for manual inspections while enhancing the safety on a construction site.

## 1. INTRODUCTION

Ensuring workers safety on a construction site is crucial, requiring regular inspection and maintenance of the structural assets to avoid accidents. Site managers play a key role making critical decisions that impact safety and operational efficiency. According to accident causation model [1], accidents can be attributed to unsafe conditions and unsafe actions. Monitoring these two causes can reduce the safety risks on a construction site. Given the dynamic working circumstances and complex nature of a construction site, workers are progressively exposed to hazardous environments. Identifying the causes of unsafe conditions, actions, inspecting equipment, conducting safety evaluations are the activities that are involved within safety and health monitoring [2]. Accidents occurring on a construction site have led to fatalities, asset damage, injuries to workers, financial loss, and delays [3].

Accidents may arise due to construction errors, insufficient protection equipment, inadequate technical specifications, excessive loads on scaffoldings, non-compliant components, improper physical actions, distraction, and various other factors [4], [5]. Factors influencing accidents on construction sites are susceptible to change from time to time [3]. Scaffolding is a critical asset on a construction site that concerns safety-related risks. Majority of the construction is performed with the use of scaffolding. It contains steel tubes and joints, and the tube axes describes the spatial configuration. It serves as a temporary structure that facilitates access to elevated areas during and post construction. The collapse of scaffolding can result in loss of life and financial costs, requiring strict compliance with safety protocols [6]. Regular inspections and required maintenance are performed to ensure good working condition and safety of the workers operating at heights and pedestrians around the scaffolding from any kind of accidents.

The site manager plays an essential role in coordinating and supervising the construction site, ensuring the safety of workers, timely project completion and budget adherence. Site managers are accountable for the inspection and monitoring of the scaffoldings. The current scaffolding inspection relies mainly on visual inspections [7] which are inefficient and may lead to errors, especially for a large and dynamic construction site. During construction, scaffolding structure often undergo modifications which may sometimes lead to deviating from the standard guidelines, necessitating regular inspections. This modification may result in missing or incorrectly replaced scaffolding components which may impact stability. Continuous monitoring of scaffolding to detect modifications and verify compliance requires time and effort from the site manager. As the construction progresses and for a complex site, manual inspection of scaffoldings becomes challenging and inaccurate [8], which may pose a potential threat to the safety of workers and individuals situated on and around the scaffolding. Comparison of the scaffolding structure with the reference structure which is according to the established design rules is required to identify any modifications. Additionally, periodic maintenance needed to restore the structural integrity.

Prognostics and Health Management (PHM) is a method that focuses on the degradation mechanism of an asset to forecast its health to enhance maintenance optimization. PHM framework ensures optimal functioning, enhanced reliability, and minimal maintenance cost of an asset [9]. It involves steps like anomaly detection, wherein an alert is generated whenever acquired data deviates from required, or desired performance. Later diagnostics is to find the probable cause for the deviation in location and nature. Finally, prognostics to estimate the time when an asset will no longer be able to perform its intended function given the current health and the degree of degradation [9]. Figure 1 shows a workflow of the proposed scaffolding inspection system, which utilizes the concepts of PHM for asset monitoring and health management. The process begins with data acquisition, where 3D point cloud data scans of scaffolding structure are collected. When a scaffolding is installed by experts, an initial scan is performed which serves as a reference or certified scan. During structure health monitoring, the recent acquired data is continuously compared with the certified or reference data to highlight the modifications in the scaffolding structure. If a structure modification is detected, that may indicate a potential safety concern, the system alerts the site manager. This workflow enables automated inspection of scaffolding which might reduce the time and effort repeatedly required by the site manager. Furthermore, the error caused by manual visual inspection can be minimized, thereby assisting site manager in their decision-making.

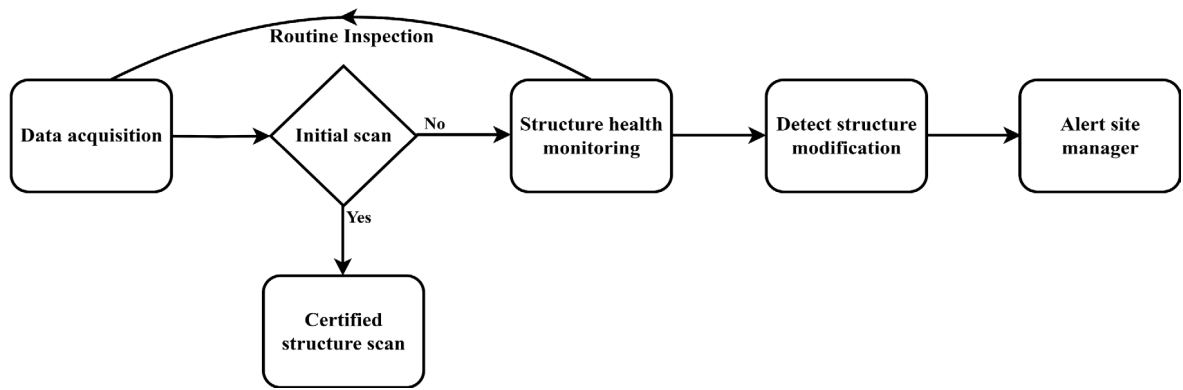

*Figure 1. Workflow of proposed scaffolding inspection system*

The key contributions of this paper are as follows:
1. A cloud-based distributed platform empowered by AI for processing, analysing and visualizing of scaffolding structure, thereby enabling automated inspection and decision making.
2. A method for monitoring scaffolding using 3D point cloud data. This facilitates the detection of structural changes that impacts its stability, hence enhancing the safety at the construction site.
3. An approach to represent the scaffolding using a graph data structure, which enables the development of a design rule engine for compliance checking.

The rest of the paper is structured as follows. Section 2 reviews the literature on traditional and current approaches for inspection of scaffolding and highlights the research gap. Subsequently, section 3 presents a methodology used for proposed approach, followed by the results in section 4. The conclusion and future directions of this work are presented in section 5.

## 2. STATE OF THE ART

### 2.1 Approaches of scaffolding inspection

Construction workers often face safety risks, and the assessment of scaffolding structures health is still premature [10]. The variability of design parameters, depending on the load conditions, the boundary conditions, and material, makes monitoring of scaffolding difficult. [10] employs wireless strain sensors to collect data from the scaffolding prototype and transmit it to a Finite Element Machine (FEM) model which assesses the real-time structural conditions. [7] explores a comprehensive approach to scaffold monitoring that integrates aspects like strain data, finite element machine, machine learning, and scaffolding. The structural state is assessed using a strain gauge sensor, and a data-driven approach is implemented to monitor the scaffolding. A finite element machine model is designed to replicate the structural response for a given load condition, and hence generating, training data for a machine learning model was generated using random loading scenarios. Finally, structural conditions were classified as safe, overloaded, unevenly settled, and risk of overturning by using a machine learning algorithm.

The inspection of scaffolding is labour-intensive and subjective, sometimes resulting in inconsistencies. Computer vision and machine learning is used to develop an automated inspection process [11]. Both real and synthetic labelled datasets are employed to train machine learning models for detection of classes like rust, cracks, and deformations on scaffolding components. Building Information Modelling (BIM) and image processing methods are combined to estimate the scaffolding progress by comparing it with site photos [12]. The safety monitoring models are restricted to specific hazards; however, the dynamic nature of the construction site makes it impractical to detect various hazards [13]. Scaffolding inspection is conducted to ensure structural stability and workers safety. Furthermore, accidents may arise due to workers action while on the scaffolding. The authors in [13] proposed an approach to detect workers unsafe actions while on a mobile scaffolding by analysing correlations drawn from safety rules. A deep learning model for object detection and instance segmentation is implemented on the manually labelled images of scaffolding with and without outriggers. The identification of workers unsafe action is achieved by integrating classification and segmentation of workers tasks with an object correlation detection module. The object correlation module determined the correlation between the predicted task and specific conditions. To mitigate the risk of falls from height in the construction industry, authors in [14] proposes a Smart Safety Hook monitoring system. This system integrates computer vision with Internet of Things monitoring technologies. Deep learning is employed to identify workers and scaffolding from live-camera feed. Upon detection of a worker in a risk zone, the Arduino Nano equipped with an inertial measurement unit and altimeter sensor will activate to check the status of the safety hook.

The aforementioned approaches are mostly implemented and evaluated on either a scaffolding prototype or a part of the scaffolding. Moreover, the models rely significantly on the quality of the collected image data, which can be affected by the lighting conditions and angles of image acquisition. The training of machine learning models requires labelled data, which can be labour-intensive, and the model's evaluation depends on the ground truth. All approaches require additional development and validation prior to their application in real-world data.

## 2.2 LiDAR based approaches

3D laser scanning or Light Detection And Ranging (LiDAR) emits laser beams and detects the reflected signals from the target to determine the distance. Scanners utilizing the time-of-flight technique emits a laser pulse with a predetermined velocity and measures the time duration of the reflected pulse. The distance to the target can then be estimated. LiDAR create a three-dimensional map of physical environment by measuring distances and spatial interactions among the objects. 3D point cloud data obtained from laser scanning technology accurately and efficiently captures spatial structures [15]. The technology provides high frame rates, extensive field of view, and the capability to operate in various lighting conditions, making it suitable for dynamic and complex construction environments [16].

3D point cloud data has emerged as an invaluable tool in construction, offering applications that improve efficiency and accuracy through progress tracking, enhancing safety throughout the project life cycle through quality inspection [17]. LiDAR technology utilized in [18], [19], [20], [21] to perform as-built inspection, quality assessment of buildings. Authors in [18] aims to derive pictorial, geometrical, spatial, topological, semantic information from architectural objects like historical buildings and monuments. [18] is a practical demonstration of generating semantic models from high resolution images, following calibration and alignment to produce point cloud data. The semantic models are subsequently integrated into computer-aided architectural design systems for advance analysis and documentation. The research in [19] presents an approach of employing high-resolution digital cameras and photogrammetry techniques to remodel building facades adversely affected by visual pollution due to urban development. [20] integrates 3D computer-aided design (CAD) with laser scanning technologies to address the issues such as progress tracking and quality control. The authors in [20] introduces an approach to optimally register CAD models with site scan objects, and subsequently computes the as-built poses, enabling the verification of dimensional compliance with specifications. The authors in [21] proposed a method to initially extract crack information from 2D images and then reconstruct it into 3D scene utilizing the structure from motion (SfM) algorithm. This method provides a blend of image processing and 3D modelling to assess structural health.

The study conducted in [22] aims to reduces the time-consuming, and error prone manual inspection of highway retaining walls by extracting geometric features from laser scan data to identify displacement in Mechanically Stabilized Earth (MSE) wall. The work in [22] extracts horizontal joints from point cloud data of the MSE wall obtained at different time intervals, which serves as a benchmark for measuring displacement. The existing approach to safety planning at excavation site susceptible to cave-ins and falls is manual and prone to error. The authors in [23] develops a semi-automated approach that utilizes the geometrical characteristics of 3D point cloud data to detect fall and cave-in hazards. To enhance safety planning, hazards identified from point cloud data and necessary safety rules are integrated with Building Information Modelling (BIM). Active safety systems which can detect hazards in real-time are essential for improving the safety on a construction site. Authors in [16] explores the use of 3D range camera for accident prevention in a construction site. Algorithms are developed to enhance the ability of 3D range cameras in detecting and tracking objects. Real-time feedback can be provided to the operator which enhances the safety during heavy machinery operation.

On a construction site, scaffolding-related issues pose a primary source of injuries and accidents. Safety features like toe-boards and guard-rails are provided to avoid falls and other related injuries. The authors in [8] presents an automated inspection method to ensure compliance with the safety regulations. Upon detecting the location of vertical element of scaffolding from point cloud data, a planar and horizonal surface is identified for scaffolding platforms. The four sides of the detected work platform give the toe-board and guard-rails. Safety regulation assessment, including the height of the toe-board, number and location of guard-rails is conducted on the extracted toe-board and guard-rails components. The focus of [8] about the scaffolding safety inspection is predominantly on the toe-board and guard-rails, rather than the complete structure. The author in [24] presents a model for 3D deformation monitoring of scaffolding. Owing to restricted scanning range of LiDAR, [24] employs point cloud data acquired by multiple stations, referred to as multi-thread LiDAR technology. The proposed model comprises of point cloud alignment and tube axes modelling. The geometric relationship of the planar features in the scan is utilized to compute the transformation parameters for point cloud alignment. For scaffolding tube axes modelling, [24] integrates least-square method with RANdom SAmple Consensus (RANSAC) algorithm. The deformation of scaffold is monitored by comparing tube axes model at various time interval. This method of point cloud alignment necessitates a sufficient planar surface in the scan. A 3D reconstruction approach for scaffolding monitoring is proposed in [25], RandLA-Net a semantic segmentation method is implemented to detect scaffolding from a point cloud data collected using robotic dog. The deep learning model is trained both from scratch and through transfer learning with the Semantic3D dataset. The 3D CAD model is generated from the predicted scaffolding point cloud. is more focused on identification and 3D reconstruction of scaffolding on the construction site. On a large construction site, direct processing of point cloud to identify small objects is computationally inefficient. The authors in [26] presents a methodology that integrates the advantages of 3D point cloud data and 2D image data to detect and locate the unsafe scaffold joints. 3D semantic segmentation is used to isolate scaffolding from acquired point cloud data. The scaffold joints are identified using coordinates of upright and guard rails and subsequently generating a joint image. A deep learning model was trained to detect the ledger end and tail from the scaffold joint images, providing insights about the safety status of the joint.

*Table 1: Previous studies on scaffolding inspection using point cloud data*

| Paper | Objective | Methodology | Data acquisition | Focus |
| --- | --- | --- | --- | --- |
| [8] | Detection of toe-boards and guard rails | Using geometrical properties of point cloud data, the four sides of work platform was extracted after detecting the uprights (vertical) and platform (horizontal) | Terrestrial Laser Scanner | Toe-boards and guard rails |
| [24] | Monitoring scaffoldings for deformations | Point cloud alignment and then comparing tube axis model for deformations | Multi thread LiDAR | Planar surface for alignment |
| [27] | Scaffolding extraction | 3D semantic segmentation model to identify scaffolding | Mobile Laser Scanner | Scaffolding identification |

| [26] | Inspection of scaffold joints | Semantic segmentation to extract scaffolding and then point-to-image translation of joints for safety inspection | Terrestrial Laser Scanner | Scaffold joints |

To the best of our knowledge, there is limited research that integrates scaffolding monitoring with point cloud data. The author in [24] indicates lack of study employing point cloud data for the deformation of temporary structure. Table 1 demonstrates the research that incorporates point cloud data for scaffolding is limited to either specific parts of the scaffold or the detection of scaffolding. Comprehensive methods for monitoring complete scaffolding structures using point cloud data are lacking. The purpose of this paper is to explore the use of AI and digital technologies into scaffolding inspection and assist site managers in enhancing construction site safety. The objective is to develop a cloud-based AI-driven platform to automate the monitoring of scaffolding structure for safety assessment using point cloud data.

## 3. METHODOLOGY

This paper focuses on identifying the modification in the scaffolding during the construction progress. As discussed in section 1, the routine manual visual inspection of scaffolding is time-consuming and may be prone to errors, especially on a complex site and due to human fatigue. The following insights are based on semi-structured interviews conducted with site manager, whose one of the responsibilities is to ensure the safety of workers on a construction site. According to the manager, scaffolding inspections are repeated every few weeks and the inspector may struggle to recall specific changes made to the reference structure. The accuracy of inspections is also influenced by the inspector's familiarity to design principles or rules. During construction, worker-structure interaction poses additional risks, as workers occasionally remove the components or braces of the scaffolding for accessibility. The most common challenges encountered by the site manager during scaffolding inspection are that the elements of the structure are not replaced, or even if they are placed, it has a chance of failure to comply the design rules. These issues can compromise the structural integrity and, eventually, the safety of the workers.

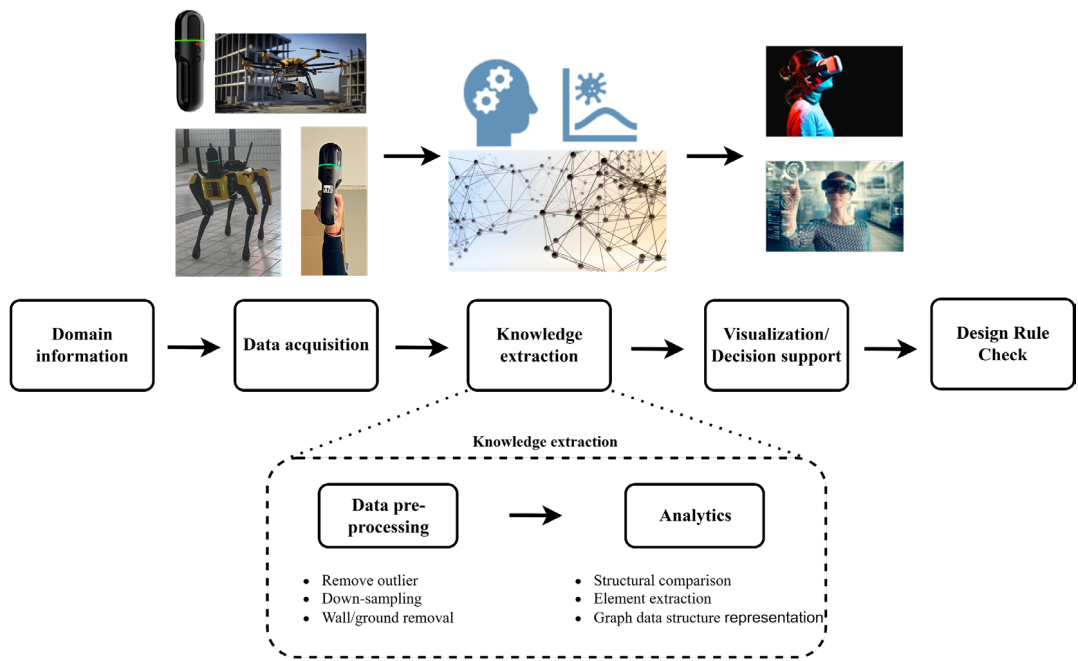

*Figure 2. AI-based cloud platform for scaffolding inspection*

This research seeks to automate the routine inspection process necessary to maintain the structural integrity of the scaffolding and ensure safe working conditions. It further explores the role of AI and digital technologies to assist site manager with informed decision-making on a construction site. As illustrated in Figure 1, the proposed scaffolding monitoring framework is structured into three parts: the acquisition of certified or the reference structure and campaign-based data acquisition, then performing a routine inspection for structure comparison and modification and finally assist site manager in decision-making. Figure 2 adapted from [28] presents the comprehensive platform used in this paper, encompassing the entire workflow from data acquisition to knowledge extraction and ultimately leading to visualization. The platform follows microservices-based architecture, enabling seamless data acquisition, secure transfer, storage, processing, analytics and visualization.

The range of various sensing devices like Light Detection and Ranging (LiDAR) scanners and robotics platforms, including autonomous robot dog and drones are employed to capture raw 3D data. The acquired data is subsequently transferred to a BLOB storage which handles the large binary files and in this case LiDAR output files. A LiDAR sensor operated by sweeping laser across a wide surface and collects millions of distance measurements, producing a dense 3D point cloud data. The point cloud data represents the spatial configuration of the environment and contains extensive information that requires further processing. As shown in Figure 2, the knowledge extraction stage encompasses various operations such as data pre-processing, structural comparison, and graph data structure representation. Through data cleaning i.e., by eliminating irrelevant data points such as outliers and isolating the object of interest namely the scaffolding structure, ensures that only high-quality information is retained for further analysis. This extracted scaffolding from the raw data still contains large number of points that must be represented effectively. Since scaffolding can be defined as the interconnection of rods and joints, this work employs graph data structure as an effective representation method. This representation is well suited to capture the relationships between nodes (scaffolding joints) and edges (scaffolding rods or braces). Furthermore, graph-based approaches not only preserve structural relationships but can enable the use of well-established methodologies in network analysis to better understand the scaffolding structure. Structural changes occurring during the construction process can affect stability. Therefore, the processed and extracted scaffolding point cloud data is compared against the certified reference scaffolding design to detect the structural changes or deviations. Hence, continuous monitoring of scaffolding is essential and any alterations in the structure can be highlighted and reported promptly. By drawing attention to these areas of modification, the site manager can prioritize the inspections on critical zones, and enable targeted corrective actions, optimizing both time and resource allocations. On this basis, the digital technology becomes a valuable tool to assist the site manager making an informed decision on the health of the scaffolding and the safety of workers. Finally, the visualization stage employs immersive technologies like Augmented Reality (AR) and Virtual Reality (VR) to present the extracted knowledge, supporting decision-making through intuitive interaction with the processed data. Prior to data acquisition, construction domain knowledge is incorporated into the platform to guide both data collection and processing strategy. Although the developed platform focuses on data acquisition, knowledge extraction and visualization, it can be further extended to incorporate the design rule check module. Such an addition would validate the extracted scaffolding model against the established design rules and regulatory requirements, thereby strengthening compliance and enhance worker safety.

The following section presents detailed results corresponding to the three stages of Figure 2, i.e., data acquisition, knowledge extraction and visualization, demonstrating the effectiveness of the proposed platform.

## 4. RESULTS AND DISCUSSION

This section demonstrates the achieved results during different stages of the developed AI-based platform.

### 4.1 Data acquisition

is shown in Figure 3 illustrates the raw point cloud data of scaffolding from a construction site which is acquired using LiDAR device. The figure presents different point of view of the acquired data. As shown in figure the raw 3D point cloud scan obtained from a construction site consists of approximately 16 million points and includes not only the scaffolding but also contains objects such as nearby trees, and other surrounding structures which are not of interest. The next section describes the processing of point cloud data to isolate the primary object of interest from the raw data.

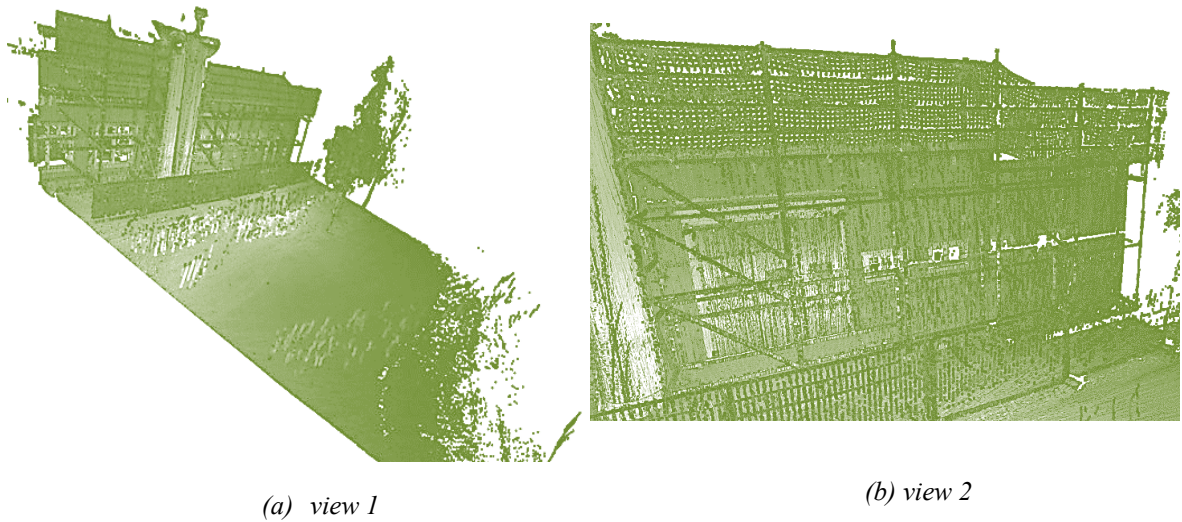

*(a) view 1*  *(b) view 2*

*Figure 3. Raw point cloud data of scaffolding*

## 4.2 Data pre-processing

The pre-processing step which is discussed in this section is applied to both the initial and periodic data acquisition. Raw data points often contain noise, outliers, and redundant points, therefore pre-processing steps ensures that the dataset is cleaner and more interpretable. To extract the object of interest, the pre-processing steps involve several operations like removal of ground plane, elimination of outliers, and voxelization to reduce the data point density and make the data manageable for further processing. In voxelization, point cloud data is partitioned into uniformly sized grid and points within each grid is aggregated to achieve a manageable resolution. For planar surface removal, iterative algorithms such as RANSAC [24], [29] which identifies and separates the planar surface from the rest of the point cloud data is used. In this study, RANSAC is employed by repeatedly fitting planes to some random subsets of data points and selecting plane with the highest number of inliers. This allows the planar surfaces to be identified and removed, leaving the scaffolding structure as the primary object of interest. Figure 4 shows the detected ground plane and wall in red. Since the scaffolding is installed at a certain defined distance from the wall, all the data points beyond that distance are eliminated, resulting in the object of interest as shown in Figure 5. In this paper, the certified scaffolding structure is considered as the initial or reference scan.

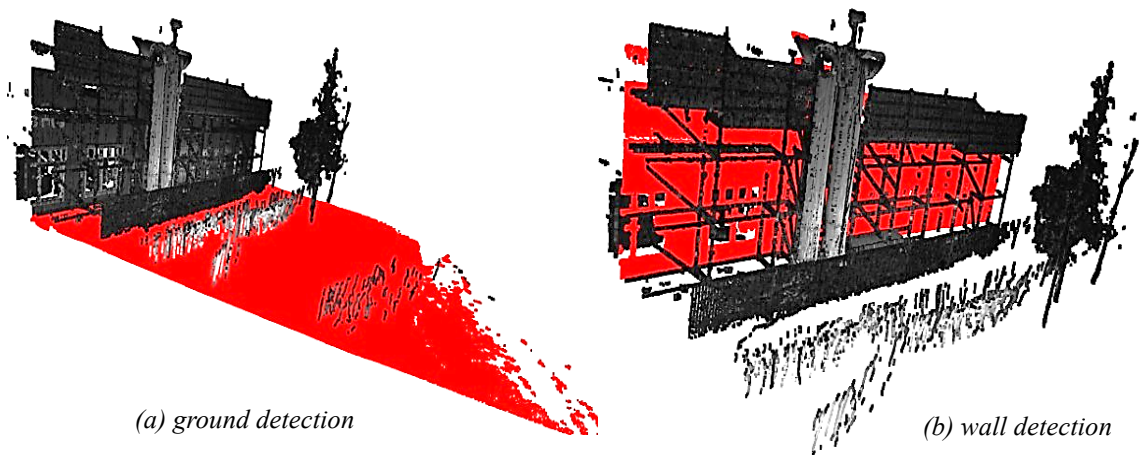

*(a) ground detection*  *(b) wall detection*

*Figure 4. Processing of point cloud data*

## 4.3 Structure comparison

This step replicates the routine inspection process typically undertaken by site manager, where in every two weeks or as per the requirement, a LiDAR scan of the same construction site is carried out. For periodic monitoring, campaign-based LiDAR scans are acquired and is subsequently compared to the certified reference scan data to detect modifications in the scaffolding structure. The new acquired point cloud data undergoes the pre-processing steps as mentioned in section 4.2. Two of the most common issues faced by the site manager during inspections are missing and deviated scaffolding braces or elements. Such changes usually arise as the construction progresses and workers modify the temporary structure for their ease of access, which is shown in Figure 6. In the figure, the missing element is shown as a red circle whereas the shifted or deviated elements are indicated by red arrows.

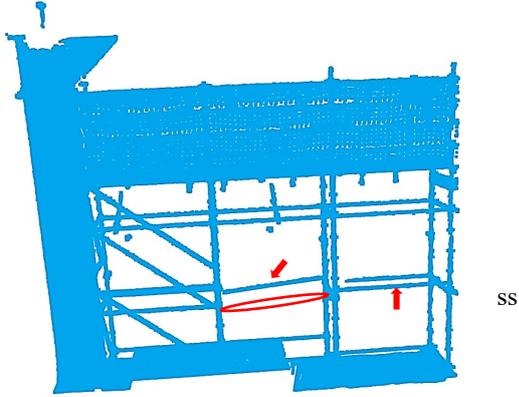

*Figure 5. Campaign based scan with modifications on the structure*

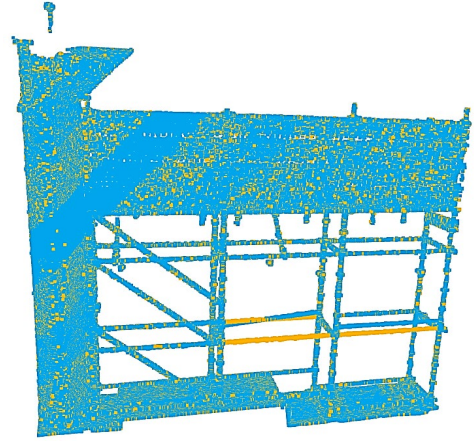

*Figure 6. Structure comparison*

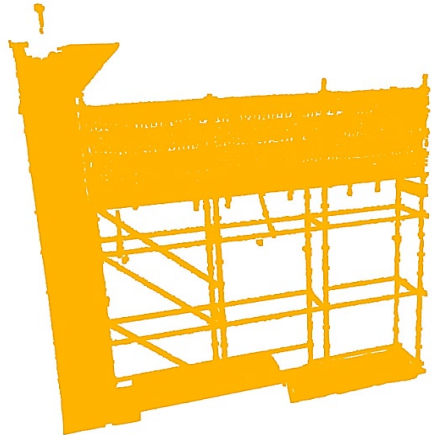

*Figure 7. Initial scaffolding point cloud data*

The recent clean extracted scaffolding point cloud data is compared with the reference certified scaffolding data to detect any modifications made, so that the site manager can focus only on those modified sections. Iterative closest point (ICP) algorithm [30] is used to align and minimize the distance between the two point cloud data. The ICP algorithm estimates the transformation matrix, a combination of rotation matrix, R and a translation matrix T, that minimizes the mean squared error as given in Equation (*1*. Here, I represent the data from certified reference scan and C the current scan, and $N_p$ denotes the total number of data points.

$$Error\ (R,T) = \frac{1}{N_p} \sum_{i=1}^{N_p} \|I_i - RC_i - T\|^2 \qquad (1)$$

For each point in C, the algorithm identifies the closest point in I to establish a point correspondence. These correspondences are then used to estimates the optimal transformation matrix that best align the two point cloud dataset. To check for convergence, the algorithm evaluates the change in error and iterate until convergence, or the maximum iteration limit is reached. Figure 7 shows the output from the algorithm, where matching or identical points between the scans appear in blue colour, while the modified points are highlighted in yellow. This provides the site manager to focus on certain sections requiring attention, which is particularly beneficial for a large complex construction site with scaffoldings at multiple locations.

To further evaluate and to visualize the alignment accuracy between the current and the reference point cloud dataset, the distance for each point data is computed. A threshold is applied to distinguish between well-aligned and poorly aligned points, with result colour-coded depending on the severity level. If the distance between current scan data points and certified scan data points are within the threshold, the points are coloured green, while those exceeding are displayed in red, signalling possible misalignment and for closer inspection. By tuning the threshold parameter in point cloud distance calculation, the percentage of deviation or modification can be adjusted. The flexibility of tuning the threshold allows site manager to adapt the severity level to site specific requirement and professional judgement. Figure 8 shows the points having distance value less than the threshold in green whereas points with larger distance then threshold in red. Figure 8(a) shows results using 10% threshold, where a small deviation is acceptable and do not compromise the structural integrity, whereas Figure 8(b) demonstrate the result with 5% threshold where even a small deviation is unacceptable.

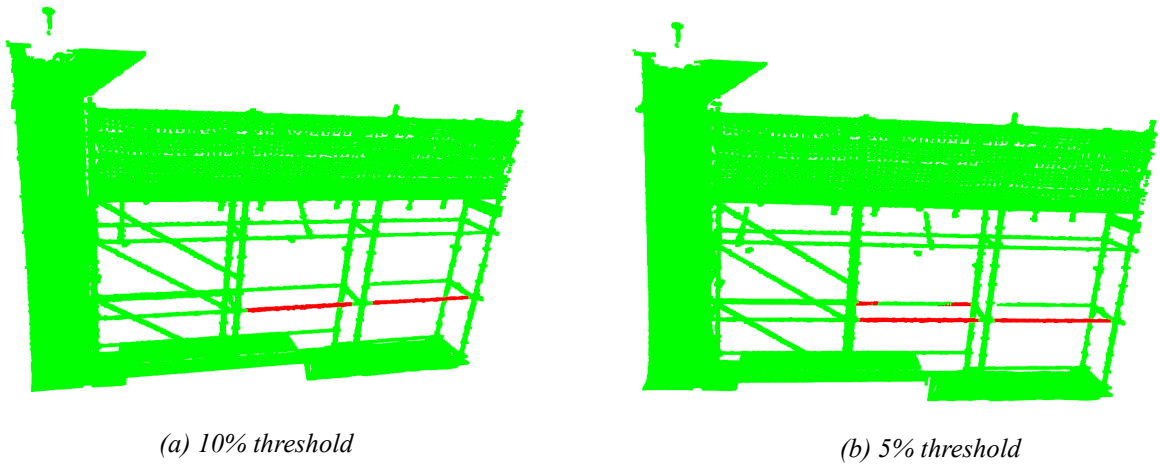

*(a) 10% threshold*            *(b) 5% threshold*

*Figure 8. Distance comparison*

## 4.4 Graph representation

Scaffolding structures, composed of interconnected elements such as tubes or braces and joints which can naturally be represented using a graph data structure. The physical elements like rods or braces of scaffolding represents edges in a graph data structure, whereas joints represent the nodes. Using this approach, it is possible to represent a complex point cloud data of scaffolding in an efficient and systematic way. Figure 9 illustrated the steps involved from raw point cloud data to graph data structure representation of a scaffolding. In the first step, scaffolding is extracted from the raw point cloud data. To segment the point cloud data according to local geometric characteristics, the KDTree algorithm [31], [32] is employed to identify neighbouring points within a pre-defined radius. Singular Value Decomposition is then applied to compute the principal components, which are used to determine the distribution of the selected points and classify their shape as linear, planar, or spherical. This classification allows braces to be identified as linear features, platform and safety sheets as planar, whereas joints as spherical features. The result of this step is shown in Figure 9 under element extraction block, where braces are represented in green, joints in red whereas planar surface such as safety sheet in blue. Subsequently, the Density-Based Spatial Clustering of Applications with Noise (DBSCAN) [33], [34] algorithm is implemented to cluster the point cloud data and isolate the linear features that corresponds to the scaffolding braces. To reduce the data and computation complexity two farthest points are determined for each cluster along with the alignment of the braces. For each of the farthest points, the corresponding close data points are identified, and their mean is computed to represent a physical joint of the scaffolding. As shown in figure, braces are represented in green, joints in red are the mean derived from bounding box which are represented in dotted lines. The last row of Figure 9 presents the colour coding for each brace in horizontal x direction, horizontal y direction and vertical direction of a scaffolding.

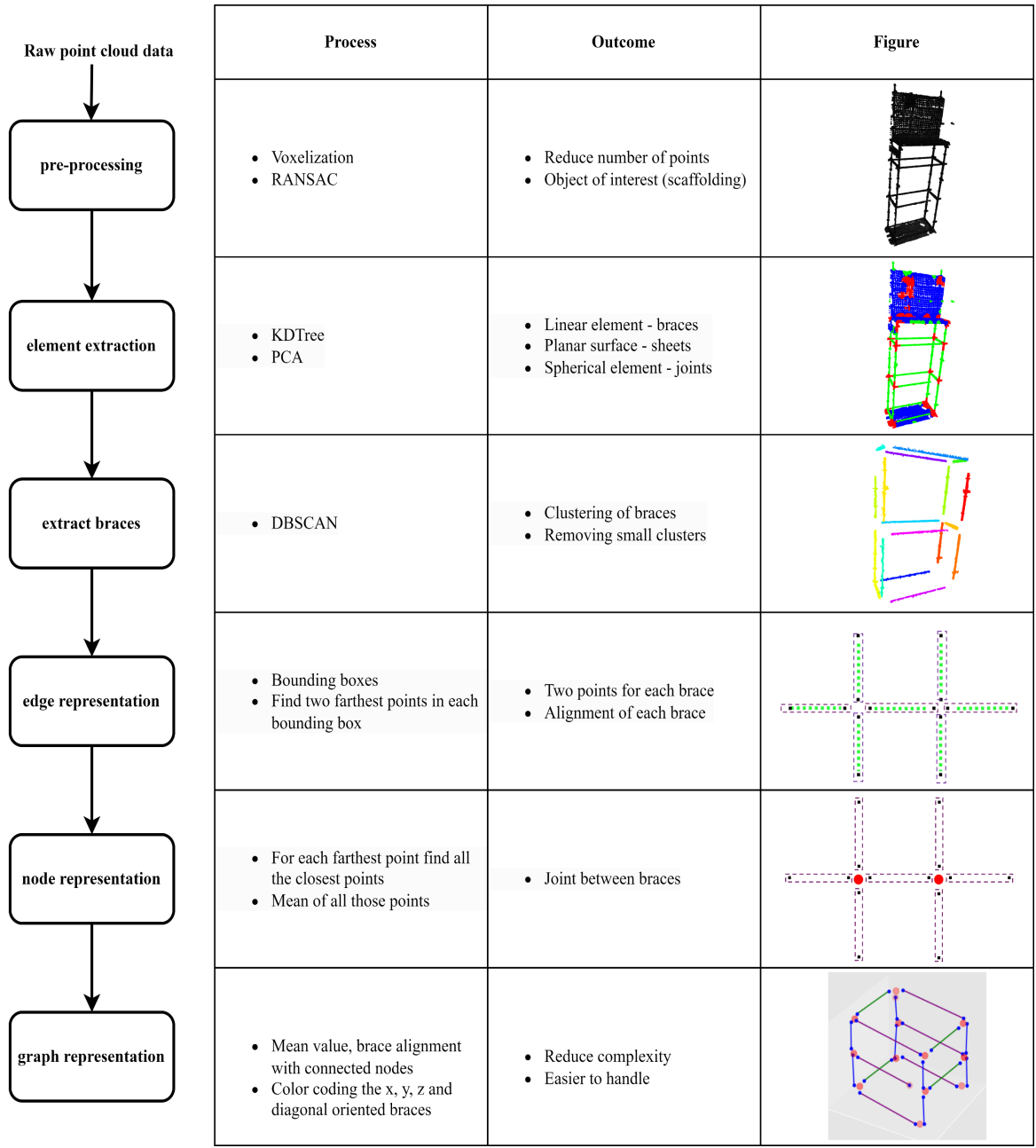

*Figure 9. Steps involved for graph data structure representation*

### 4.4.1 Hybrid clustering

In the element extraction block, linear elements i.e., braces are separated from spherical elements like joints, after which the DBSCAN algorithm is applied to cluster the braces. However, as shown in Figure 10(a), certain clusters contains both vertical and horizontal braces under one label. This edge cases requires further refinement to cluster. To address these edge cases, an additional normal vector-based separation is introduced, specifically targeting the wrong or misclassified clusters. The normal vector representation for edge case is shown in Figure 10(b). Accordingly, for edge cases two-stage clustering or hybrid clustering is adopted, the first stage is based on spatial distance, while the second stage employs normal-vector direction. As shown in Figure 10(c), the results generated from both the clustering stages are combined, this enables a more accurate differentiation between vertical and horizontal braces within the scaffolding structure.

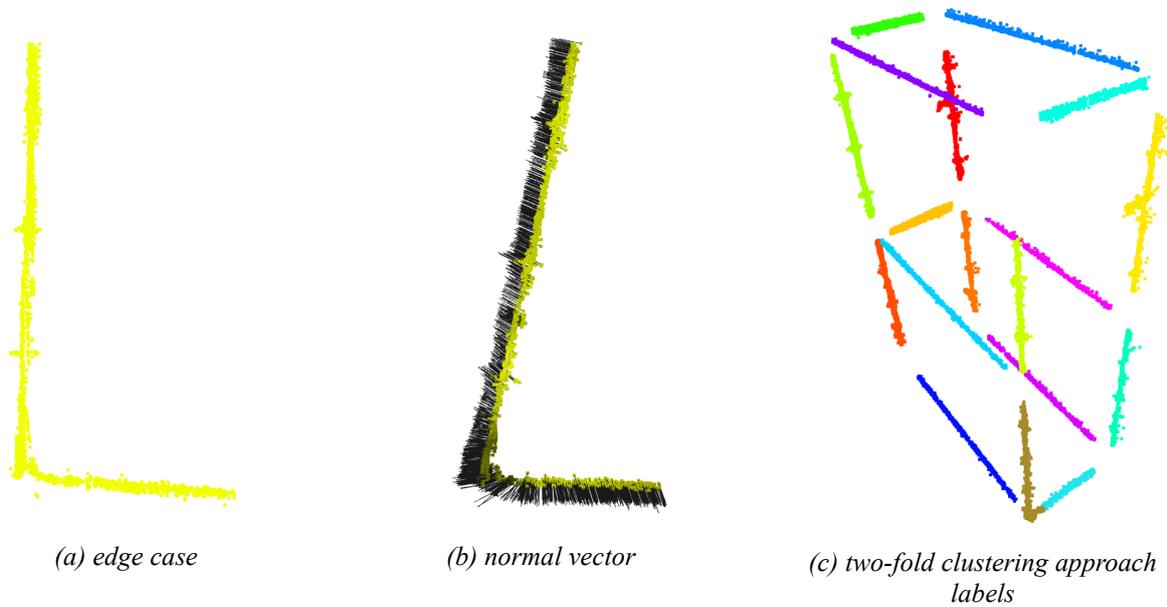

*(a) edge case*  *(b) normal vector*  *(c) two-fold clustering approach labels*

*Figure 10. Hybrid clustering*

## 4.5 Visualization

Once the scaffolding structure is represented as a graph data structure, its layout is shown in Figure 11. In this representation, horizontal braces are coloured green whereas vertical braces are shown blue colour, and joints are denoted as small red circles. Although, Figure 11 may not visually resemble a physical scaffolding structure, it effectively captures the connectivity and relationships between components. As discussed previously, one of the major issues during inspection is the missing scaffolding elements. Figure 12 demonstrates this by comparing the graph representation of certified reference scan with the recent scan in which a brace is missing. Figure 12 shows the graph representation of a scaffolding with missing brace highlighted in red.

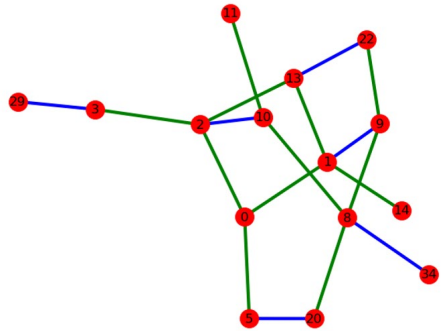

*Figure 11. Graph data structure representation of scaffolding. Horizontal braces are shown in green, vertical braces as blue and joints as red circles*

Virtual Reality (VR) and Augmented Reality (AR) are increasingly recognized as transformative technologies that are changing construction industry through better visualization capabilities [35]. VR creates an experience by immersing users in a fully simulated environment, effectively isolating them from the physical world. This allows users a natural interaction and navigation within the virtual environments in a manner that closely resembles reality. On the other hand, AR superimposes digital information directly onto the physical environment, enhancing the user's interaction with their surroundings. These immersive tools enables site manager to conduct virtual inspection of scaffolding and specifically focus attention on critical sections to ensure structural integrity. A screenshot from the AR glasses is shown in Figure 13, where missing brace is represented in red colour and deviated element of the scaffolding in yellow colour. These technologies break down the geographical barriers, supporting problem-solving and coordination, extending collaboration practices by diving into 3D models and perform hands-on that goes beyond computer screens. This, in turn, promotes real-time discussions among the experts, resulting in improved decision-making.

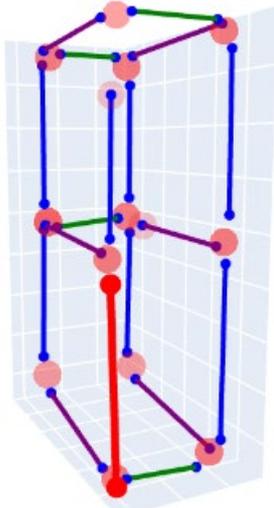

*Figure 12. Comparison of scaffolding structure using graph representation*

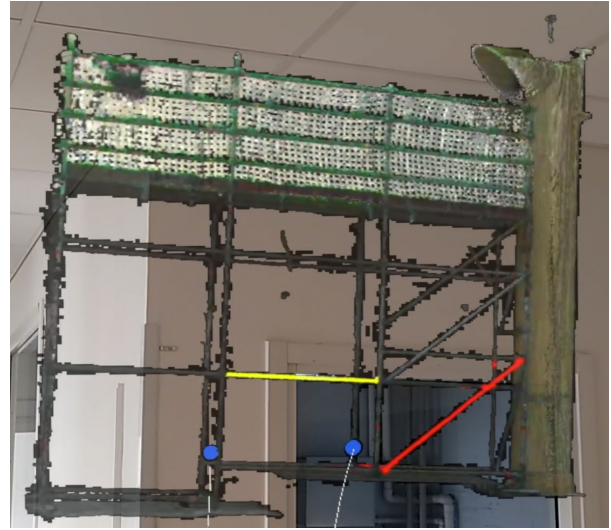

*Figure 13. Site manager visual using AR glasses*

## 5. CONCLUSION AND FUTURE WORK

This paper proposed a cloud-based AI platform for monitoring scaffolding on construction site. The method processes the raw point cloud data acquired using 360-degree LiDAR scan, to isolate the scaffolding. Then compares the certified reference scaffolding structure scan with the recent scan taken at regular interval. Deviations or missing elements of a scaffolding are detected through alignment algorithms. To quantify the difference between the reference and recent scan, point cloud distance is computed enabling precise identification of structural modifications. By automating the scaffolding monitoring on complex construction site, the proposed approach reduces the time and effort required from the site manager. Additionally, this paper also suggests an efficient way to represent the scaffolding using a graph data structure, which not only captures the connectivity of joints and braces but also offers to integrate numerous well-established graph algorithms for advance analysis. Highlighting the structural modifications during construction progress directs the site manager attention to focus on that critical scaffolding section, thereby reducing human errors during inspection. By providing data-driven insights into scaffolding structural modifications, the proposed approach assists site manager in decision-making. Moreover, the integration of advance visualization tools like VR/AR further enhances remote visualization, foster real-time discussion and facilitates informed decision-making. The proposed cloud-based platform demonstrates the use of AI and digitization to automate scaffolding inspections which currently rely heavily on visual inspections. Such inspections can be time consuming, error prone and inconsistent, potentially compromising safety on a construction site. The proposed approach is one of the ways to enhance the safety of the workers and people on a construction site by reducing errors associated with the current visual inspection.

To increase construction site safety, early detection of wear, damage or unintended changes is critical. This work aligns with PHM concepts by enabling continuous monitoring of scaffolding to track changes or modifications against the certified reference structure, thereby supporting in predicting potential failures or safety hazards. Continuous monitoring the health of scaffolding also assists in forecasting when the elements may require maintenance, enhancing overall safety and reliability. This work establishes the foundation for future advances in scaffolding condition monitoring. When the scaffoldings are installed, adherence to design rules ensures structural stability and integrity. One promising extension is the integration of design rule verification, enabling the system to automatically assess the current condition of the scaffolding against predetermined safety specifications. Furthermore, implementing an expert system framework could advance the platform towards a higher degree of autonomous decision support system.

## ACKNOWLEDGMENT

The authors would like to acknowledge NCC, HÖ Allbygg, Byggföretagen, SBUF, Smart Built Environment and Formas for their continuous support and contributions. This work is a part of a project, and the corresponding report [36] has been submitted. During the preparation of this work the author(s) used ChatGPT, a language model developed by OpenAI to assist with grammar corrections and text refining. After using this tool, the author(s) reviewed and edited the content as needed and take(s) full responsibility for the content of the publication.

## AUTHOR CONTRIBUTION

Sameer Prabhu, Investigation; Methodology; Conceptualization; Data curation; Software; Validation; Role/ Writing - original draft.

Amit Patwardhan, Conceptualization; Data curation; Formal analysis; Investigation; Role/Writing - review & editing.

Ramin Karim, Formal analysis; Investigation; Resources; Supervision; Validation; Role/ Writing - review & editing.

**DECLARATION OF COMPETING INTEREST** - The authors have no known competing interest that can influence the outcomes of the present work.

**FUNDING SOURCE** – The present work did not receive any funding.

**DATA AVAILABILITY** – Due to confidentiality issues the data cannot be shared at the moment.